\documentclass{llncs}
\usepackage{rotating}
\usepackage{subfigure}
\usepackage{amsmath,amssymb,amsfonts}
\usepackage{algorithmic}
\usepackage{algorithm}
\usepackage{graphicx}
\usepackage{textcomp}
\usepackage{xcolor}
\usepackage{tabularray}
\usepackage{multirow}
\usepackage{makecell} 
\usepackage{pdflscape}
\usepackage{marvosym}

\begin{document}
\mainmatter              % start of a contribution
\title{Integrating Knowledge Distillation Methods:\\ A Sequential Multi-Stage Framework}
\titlerunning{Sequential Multi-Stage Knowledge Distillation}  % abbreviated title (for running head)
\author{Yinxi~Tian\inst{1} \and Changwu~Huang\inst{1}\textsuperscript{(\Letter)} \and Ke~Tang\inst{1}\textsuperscript{(\Letter)} \and Xin~Yao\inst{2}}
\authorrunning{Yinxi~Tian et al.} % abbreviated author list (for running head)
%\tocauthor{Ivar Ekeland, Roger Temam, Jeffrey Dean, David Grove, Craig Chambers, Kim B. Bruce, and Elisa Bertino}
%
\institute{Department of Computer Science and Engineering, Southern University of Science and Technology, Shenzhen 518055, China.\\
\email{huangcw3@sustech.edu.cn, tangk3@sustech.edu.cn}
\and School of Data Science, Lingnan University, Hong Kong, China.}

\maketitle              % typeset the title of the contribution

\begin{abstract}
%Knowledge distillation (KD) transfers knowledge from large teacher models to compact student models, enabling efficient deployment on resource-constrained devices. While diverse KD methods—including response-based, feature-based, and relation-based approaches—capture different aspects of teacher knowledge, integrating multiple methods or knowledge sources has become promising for improving student performance. However, existing multi-source integration strategies often suffer from complex implementation, inflexible method combinations, and catastrophic forgetting, limiting their practical effectiveness.
Knowledge distillation (KD) transfers knowledge from large teacher models to compact student models, enabling efficient deployment on resource-constrained devices. While diverse KD methods—including response-based, feature-based, and relation-based approaches—capture different aspects of teacher knowledge, integrating multiple methods or knowledge sources is promising but often hampered by complex implementation, inflexible combinations, and catastrophic forgetting, limiting practical effectiveness.
%This work proposes SMSKD (Sequential Multi-Stage Knowledge Distillation), a flexible framework that integrates heterogeneous distillation methods through sequential manner. Unlike prior approaches combining multiple objectives simultaneously, SMSKD adopts a progressive training strategy: at each stage, the student is trained with a specific distillation method, while a frozen reference model from the previous stage serves as an anchor to preserve acquired knowledge, thereby mitigating catastrophic forgetting. Furthermore, we introduce an adaptive weighting mechanism based on the teacher's true class probability (TCP) that dynamically adjusts reference loss strength for each sample, balancing knowledge retention and integration. By design, SMSKD supports arbitrary combinations and numbers of distillation methods, and introduces negligible computational overhead.
This work proposes SMSKD (Sequential Multi-Stage Knowledge Distillation), a flexible framework that sequentially integrates heterogeneous KD methods. At each stage, the student is trained with a specific distillation method, while a frozen reference model from the previous stage anchors learned knowledge to mitigate forgetting. Additionally, we introduce an adaptive weighting mechanism based on the teacher's true class probability (TCP) that dynamically adjusts the reference loss per sample to balance knowledge retention and integration. By design, SMSKD supports arbitrary method combinations and stage counts with negligible computational overhead.
%Extensive experiments demonstrate that SMSKD consistently improves student accuracy across diverse teacher–student architectures and method combinations, achieving superior performance compared to existing methods. Ablation studies confirm that stage-wise distillation and reference model supervision are primary contributors to performance gains, while TCP-based adaptive weighting provides complementary benefits. Overall, SMSKD offers a flexible and resource-efficient solution for integrating heterogeneous KD methods with significant performance enhancement. 
Extensive experiments show that SMSKD consistently improves student accuracy across diverse teacher–student architectures and method combinations, outperforming existing baselines. Ablation studies confirm that stage-wise distillation and reference model supervision are primary contributors to performance gains, with TCP‑based adaptive weighting providing complementary benefits. Overall, SMSKD is a practical and resource‑efficient solution for integrating heterogeneous KD methods.

% We would like to encourage you to list your keywords within
% the abstract section using the \keywords{...} command.
\end{abstract}

\section{Introduction}
\label{sec:intro}

%Deep neural networks (DNNs) have driven significant advances in artificial intelligence, with state-of-the-art performance achieved in domains such as computer vision, natural language processing, and speech recognition~\cite{lecun2015deep}. However, the impressive performance of modern DNNs is often accompanied by substantial increases in model complexity, leading to excessive computational and memory demands. Such complexity poses significant obstacles to real-world deployment, especially on resource-constrained platforms like mobile devices, IoT sensors, and edge computing environments~\cite{cheng2018model}. To address this challenge, knowledge distillation (KD) has emerged as a powerful model compression and acceleration technique that transfers knowledge from large, complex models to smaller, more efficient ones~\cite{mansourian2025comprehensivesurveyknowledgedistillation,wang2021knowledge}.
Deep neural networks (DNNs) have driven major advances in artificial intelligence, achieving state-of-the-art performance in domains such as computer vision, natural language processing, and speech recognition~\cite{lecun2015deep}. However, these performance gains are often accompanied by substantial increases in model complexity, resulting in heavy computational and memory demands. This creates significant obstacles to real-world deployment, especially on resource-constrained platforms like mobile devices, IoT sensors, and edge computing environments~\cite{cheng2018model}. To address this challenge, knowledge distillation (KD) has emerged as a powerful technique for model compression and acceleration~\cite{mansourian2025comprehensivesurveyknowledgedistillation}.

The fundamental idea of KD is to transfer the knowledge of a large, well-trained model, referred to as the teacher model, to a lightweight model, known as the student model~\cite{hinton2015distilling,gou2021knowledge}.  
This approach enables the student model to achieve significant size reduction while maintaining competitive or even superior performance compared to the teacher. Since the pioneering work of Hinton \textit{et al.}~\cite{hinton2015distilling}, KD has evolved rapidly, resulting in a diverse array of distillation algorithms tailored to different forms of transferable knowledge. Generally, KD methods can be classified based on the number of knowledge sources involved~\cite{song2022spot}: single-source KD and multi-source KD. For single-source KD, methods can be further categorized into three major paradigms according to the type of knowledge transferred~\cite{gou2021knowledge}: (1) Response-based KD, which encourages the student to mimic the teacher's output distributions or final predictions~\cite{hinton2015distilling,cui2024decoupled}; (2) Feature-based KD, which aims to align the intermediate feature representations between teacher and student models~\cite{romero2015fitnets,miles2024vkd}; and (3) Relation-based KD, which encourages the student to learn structural relationships and interactions among data samples or feature vectors from the teacher~\cite{park2019relational}. %Each paradigm provides distinct mechanisms for effectively transferring knowledge from large teacher models to smaller student models.

%Despite the proliferation of single-source KD methods, each approach exhibits distinct behaviors and strengths, often capturing different aspects of the teacher's knowledge. 
Recent studies have begun to investigate the similarities and differences among existing distillation approaches~\cite{OjhaLRLL23,zheng2023rotatedLD}. For instance, Ojha \textit{et al.}~\cite{OjhaLRLL23} systematically examined three representative methods (\cite{romero2015fitnets,hinton2015distilling,tian2020contrastive}) across four scenarios to evaluate knowledge inheritance capabilities. Similarly, Zheng \textit{et al.}~\cite{zheng2023rotatedLD} compared response-based and feature-based distillation in object detection, revealing that even when accuracies are comparable, the learning behaviors between different methods can differ substantially. These findings suggest that different distillation methods may capture complementary aspects of the teacher's knowledge, motivating researchers to explore the integration of multiple knowledge sources in KD to leverage diverse forms of teacher information for more effective student training\cite{OjhaLRLL23,zheng2023rotatedLD}. For example, some works combine response-based and feature-based distillation losses, or fuse information from different teacher layers~\cite{zagoruyko2017paying,romero2015fitnets}. Although multi-source approaches often yield performance improvements, simply combining different knowledge sources does not guarantee optimal results, as the interactions between knowledge sources are often complex and may lead to suboptimal results without careful design and balancing~\cite{huang2024harmonizingknowledgetransferneural}.

However, existing multi-source or hybrid approaches suffer from several limitations. First, many hybrid methods impose constraints on the types of knowledge that can be combined, such as restricting integration to different layers or only allowing certain knowledge categories to be integrated~\cite{tian2020contrastive,song2022spot}. This limits flexibility and may prevent the effective combination of, for instance, feature-based and relation-based knowledge. Second, current integration techniques often require sophisticated modifications to the distillation algorithms, increasing implementation complexity~\cite{song2022spot}. Simply summing different distillation losses can be suboptimal, as the scales of these losses may differ significantly, potentially leading to imbalanced optimization and degraded performance~\cite{huang2024harmonizingknowledgetransferneural}. Moreover, the lack of mechanisms to preserve or balance knowledge acquired through different sources during the training of student model can result in catastrophic forgetting, further reducing the effectiveness of hybrid distillation approaches.

To address these challenges, this work proposes a Sequential Multi-Stage Knowledge Distillation (SMSKD) framework that flexibly integrates multiple distillation methods in a stage-wise manner. In our framework, the student model is trained sequentially over multiple stages, with each stage employing a potentially different distillation method, thereby leveraging multiple distillation methods collaboratively. Crucially, to mitigate the issue of knowledge forgetting when switching between distillation methods, we introduce a frozen reference model—copied from the student at the end of each stage—that serves as an anchor for the subsequent stage. During later stages, a reference loss term is added to the loss function of the distillation method adopted in the current stage to encourage the student model to retain the knowledge acquired in previous stages, thereby alleviating catastrophic forgetting and enabling synergistic integration of diverse distillation techniques. This design allows the framework to harness the strengths of various distillation methods while maintaining a stable learning trajectory for the student model. Comprehensive experiments across multiple teacher-student configurations and datasets are conducted to validate the effectiveness of our proposed framework. Results demonstrate that our approach consistently outperforms single-source distillation and existing multi-source approaches, achieving superior student model performance.

%Our contributions are as follows. First, we propose SMSKD, a flexible and general framework that sequentially integrates multiple distillation methods and explicitly mitigates knowledge forgetting via a reference-model mechanism, enabling collaborative integration of diverse KD strategies. Second, the framework is compatible with a wide range of distillation algorithms and supports arbitrary method combinations and stage counts, greatly enhancing applicability and flexibility. Third, extensive experiments across diverse teacher–student configurations and datasets demonstrate superior student performance compared to single-source and hybrid baselines.

%Our contributions are summarized as follows:
% \begin{enumerate}
%     \item This work proposes a flexible and general SMSKD framework for integrating multiple distillation methods, with explicitly addresses the problem of knowledge forgetting via a reference model mechanism. The framework enables collaborative distillation by sequentially applying different KD strategies while preserving previously acquired knowledge.
%     \item Our approach is compatible with a wide range of existing distillation algorithms and allows arbitrary choices of distillation methods and stage numbers, greatly enhancing applicability and flexibility.
%     \item Extensive experiments on various teacher-student configurations and datasets demonstrate the effectiveness of our framework, yielding superior student performance compared to single-source distillation and existing multi-source approaches.
% \end{enumerate}

The remainder of this paper is organized as follows: Section~\ref{sec:related_work} reviews related work on single-source and multiple-source KD methods. Section~\ref{sec:SMSKD} presents our proposed sequential multi-stage framework for integrating distillation methods. Section~\ref{sec:exp_study} provides comprehensive experimental studies to validate the effectiveness of the proposed method. Finally, Section~\ref{sec:conclusion} briefly concludes the paper.

\section{Background}
\label{sec:related_work}
This section reviews various existing distillation methods, which lay the foundation for our work. First, several single-source distillation methods are introduced including response-based, feature-based, and relation-based distillation methods. Then, some recent work on multi-source distillation is presented.

\subsection{Notation}

This work focuses on supervised classification tasks, which aim to learn a model or classifier that maps the input space $\mathcal{X} \subset \mathbb{R}^d$ to the label space $\mathcal{Y}$, where $\left|\mathcal{Y}\right| = K$ denotes the number of classes. Let $f: \mathcal{X} \to \mathcal{Y}$ denote a model trained on a dataset $\mathcal{D} = \left\{ (\mathbf{x}_i, y_i) \right\}_{i=1}^{N}$, where $\mathbf{x}_i \in \mathcal{X}$ and $y_i \in \mathcal{Y}$. Given an input feature vector $\mathbf{x} \in \mathcal{X}$, the model produces a predictive label $\hat{y}$. Typically, a DNN model $f$ outputs a vector of logits $\mathbf{z} = [z_1, \ldots, z_K]$ for the input vector $\mathbf{x}$, which are then converted into class probabilities $\mathbf{p}=\left [ p_1, \cdots , p_K \right ] $ via the softmax function: $ p_k=\mathrm{exp} \left ( z_k/\tau \right ) /  {\textstyle \sum_{j=1}^{K}\mathrm{exp} \left ( z_j/\tau \right )}$ for $k=1,\dots, K$, where $\tau$ is a temperature parameter (usually set to 1). The predicted label is typically determined as the class with the highest probability, i.e., $\hat{y} = \arg\max_{k} \left \{ p_k \right \} $. 

For clarity, throughout this work, we use $f^T$ to denote the teacher model. For a given input feature vector $\mathbf{x}$, we denote its output logits, predictive class probabilities, and the output of its $i$-th intermediate feature layer as $\mathbf{z}^T$, $\mathbf{p}^T$, and $\mathbf{h}_i^T$, respectively. Similarly, the student model and its corresponding outputs are denoted as $f^S$, $\mathbf{z}^S$, $\mathbf{p}^S$, and $\mathbf{h}_i^S$. Given a training dataset $\mathcal{D}$ and a teacher model $f^T$ pre-trained on $\mathcal{D}$, the objective of KD is to train a smaller and more efficient student model $f^S$ under the guidance of the teacher model. The aim is for the student to inherit the performance and generalization capability of the teacher while achieving reductions in model size and computational cost.

\subsection{Single-Source Knowledge Distillation}

Knowledge distillation aims to transfer knowledge from a large teacher model to a compact student model. Existing single-source distillation methods can be broadly categorized into three types according to the knowledge they exploit: response-based, feature-based, and relation-based~\cite{gou2021knowledge}.

\subsubsection{Response-based distillation methods}
Response-based distillation represents the earliest and most straightforward approach in KD, focusing on aligning the predicted logits or probabilities of the student model with those of the teacher. The vanilla KD method proposed by Hinton et al.~\cite{hinton2015distilling} is the most widely adopted response-based distillation technique, which matches the soft targets or probability distributions over classes produced by the teacher and student models via KL divergence. The distillation loss for the vanilla KD method~\cite{hinton2015distilling} is formulated as follows:
\begin{equation}
\label{eq:vanillaKD}
\mathcal{L}_{\text{KD}}(\mathbf{p}^T, \mathbf{p}^S) = \tau^2 \cdot \mathrm{KL}\!\left( 
    \mathbf{p}^T \;\|\; \mathbf{p}^S
\right).
\end{equation}

More recently, the decoupled KD (DKD)~\cite{zhao2022decoupled} separates contributions from target and non-target classes, allowing more flexible control: 
\begin{equation}
\label{eq:DKD}
\mathcal{L}_{\text{DKD}}(\mathbf{p}^T, \mathbf{p}^S) 
= \alpha \, \mathcal{L}_{\text{TCKD}} + \beta \, \mathcal{L}_{\text{NCKD}},
\end{equation}
with
\begin{equation}
\mathcal{L}_{\text{TCKD}} = \tau^2 \cdot \mathrm{KL}\!\big( p^T_{y} \,\|\, p^S_{y} \big), 
\mathcal{L}_{\text{NCKD}} = \tau^2 \cdot \mathrm{KL}\!\big( \mathbf{p}^T_{\setminus y} \,\|\, \mathbf{p}^S_{\setminus y} \big),
\end{equation}
where $p^T_{y}$ and $p^S_{y}$ denote the teacher’s and student’s probabilities for the ground-truth class, and $\mathbf{p}^T_{\setminus y}$ and $\mathbf{p}^S_{\setminus y}$ represent the distributions over remaining classes. The key advantage of DKD is that it allows independent weighting of these two components via $\alpha$ and $\beta$, in contrast to vanilla KD, in which they are coupled.

Overall, response-based methods primarily focus on aligning the output distributions of the teacher and student models, thereby confining knowledge transfer to the output layer. As a result, these approaches may overlook valuable information encoded in intermediate representations or model features.

\subsubsection{Feature-based distillation methods}

Feature-based distillation transfers knowledge through intermediate representations rather than solely relying on final predictions. FitNets~\cite{romero2015fitnets} aligns hidden activations to guide the student. Attention transfer (AT)~\cite{zagoruyko2017paying} emphasizes spatial importance, and variational information distillation (VID)~\cite{ahn2019variational} models features probabilistically to capture richer information. The core idea can be expressed generally as:
\begin{equation}
    \mathcal{L}_{\text{feature}}
    = \mathcal{D}_{\text{feat}}
    \big(
        \mathcal{F} \mathbf{(h^T)},\,
        \mathcal{F} \mathbf{(h^S)}
    \big),
\end{equation}
where $\mathcal{F}(\cdot)$ denotes the feature transformation used by a particular method, and $\mathcal{D}_{\text{feat}}$ measures 
the discrepancy between transformed teacher and student features (e.g., 
$\ell_2$ distance, attention map matching, or KL divergence over feature 
distributions). Feature-based methods guide the student to imitate the teacher’s intermediate representations, improving feature-space alignment and enhancing representation quality. These signals complement response-based approaches by providing information unavailable from final predictions alone.

\subsubsection{Relation-based distillation methods}

Relation-based distillation focuses on preserving the structural relationships 
among samples, rather than aligning individual representations. These methods 
characterize the teacher’s feature space through pairwise or higher-order 
relations and encourage the student to reproduce the same relational geometry. 
For example, relational knowledge distillation (RKD)~\cite{park2019relational} aligns pairwise distances and angles, 
probabilistic knowledge transfer (PKT)~\cite{passalis2020probabilistic} matches similarity distributions, 
correlation congruence (CC)~\cite{peng2019correlation} preserves feature correlations, and 
contrastive representation distillation  (CRD)~\cite{tian2020contrastive} employs a contrastive objective over instance 
pairs in a projected embedding space. Despite their methodological diversity, 
these approaches share a common objective: transferring the relational structure 
encoded in the teacher’s representations. This can be abstracted as
\begin{equation}
    \mathcal{L}_{\text{relation}}
    = \mathcal{D}_{\text{struct}}
    \big(
        \mathcal{R}(\mathbf{h}^T),\,
        \mathcal{R}(\mathbf{h}^S)
    \big),
\end{equation}
where $\mathcal{R}(\cdot)$ extracts structural information—such as distances, 
similarities, correlations, or contrastive embeddings, and 
$\mathcal{D}_{\text{struct}}$ measures the discrepancy between the teacher’s and 
student’s relational structures.
\subsection{Multi-Source Knowledge Distillation}

While most distillation methods are developed as standalone techniques, recent work has begun exploring how to leverage multiple knowledge sources. Multi-teacher distillation~\cite{shen2019customizing} represents one such direction, where multiple pre-trained teachers provide diverse knowledge to a single student. Each teacher may specialize in different aspects of the task or employ different architectures, enabling the student to acquire richer and more comprehensive information. However, this approach fundamentally differs from our work: multi-teacher distillation requires multiple pre-trained models, whereas our framework uses a single teacher with multiple distillation strategies to extract different facets of knowledge.

Recently, Song \textit{et al.}~\cite{song2022spot} proposed the spot-adaptive knowledge distillation (SAKD) method, which integrates teacher and student models into a multi-path routing network. The method employs adaptation layers to align intermediate or penultimate representations and uses a policy network to make per-sample routing decisions at designated branch spots. This enables selective knowledge transfer from the most informative locations within the teacher network. 

However, SAKD faces critical limitations. First, it is limited to combining two methods from different sources, one from intermediate layers and the other from the output layer. It cannot fuse multiple methods or two methods from the same source, which restricts the diversity of possible combinations. Second, as noted by Huang et al.~\cite{huang2024harmonizingknowledgetransferneural}, feature-based and response-based distillation possess inherently different learning objectives and gradient characteristics. Directly combining them without proper adaptation can lead to optimization conflicts and suboptimal performance. 
These limitations motivate our SMSKD, which tackles the challenges of combining multiple distillation methods via a stage-wise multi-source approach with careful design of complementary knowledge integration.

%\section{Sequential Multi-Stage Framework for Integrating KD Methods}
\section{Sequential Multi-Stage Knowledge Distillation}
\label{sec:SMSKD}

This work proposes a novel Sequential Multi-Stage Knowledge Distillation (SMSKD) framework that enables flexible and effective integration of multiple KD techniques in a stage-wise manner. Unlike existing KD approaches that apply a single knowledge source or combine multiple sources simultaneously to train a student model, SMSKD decomposes the training of the student model into a series of sequential stages. Each stage employs a potentially different KD method, allowing the student to progressively assimilate diverse types of knowledge from the teacher model. To address the challenge of catastrophic forgetting when switching KD methods, SMSKD introduces a frozen reference model at each stage transition, which serves as an anchor for knowledge retention. Additionally, a confidence-based weighting mechanism is employed to adaptively modulate the influence of the reference supervision, ensuring robust and reliable knowledge transfer across stages. This design allows the student model to progressively accumulate and integrate complementary knowledge from diverse KD techniques while maintaining a stable learning trajectory.

%\vspace{-0.5cm}
\begin{figure}[h]
	\centering
	\includegraphics[width=1.0\textwidth]{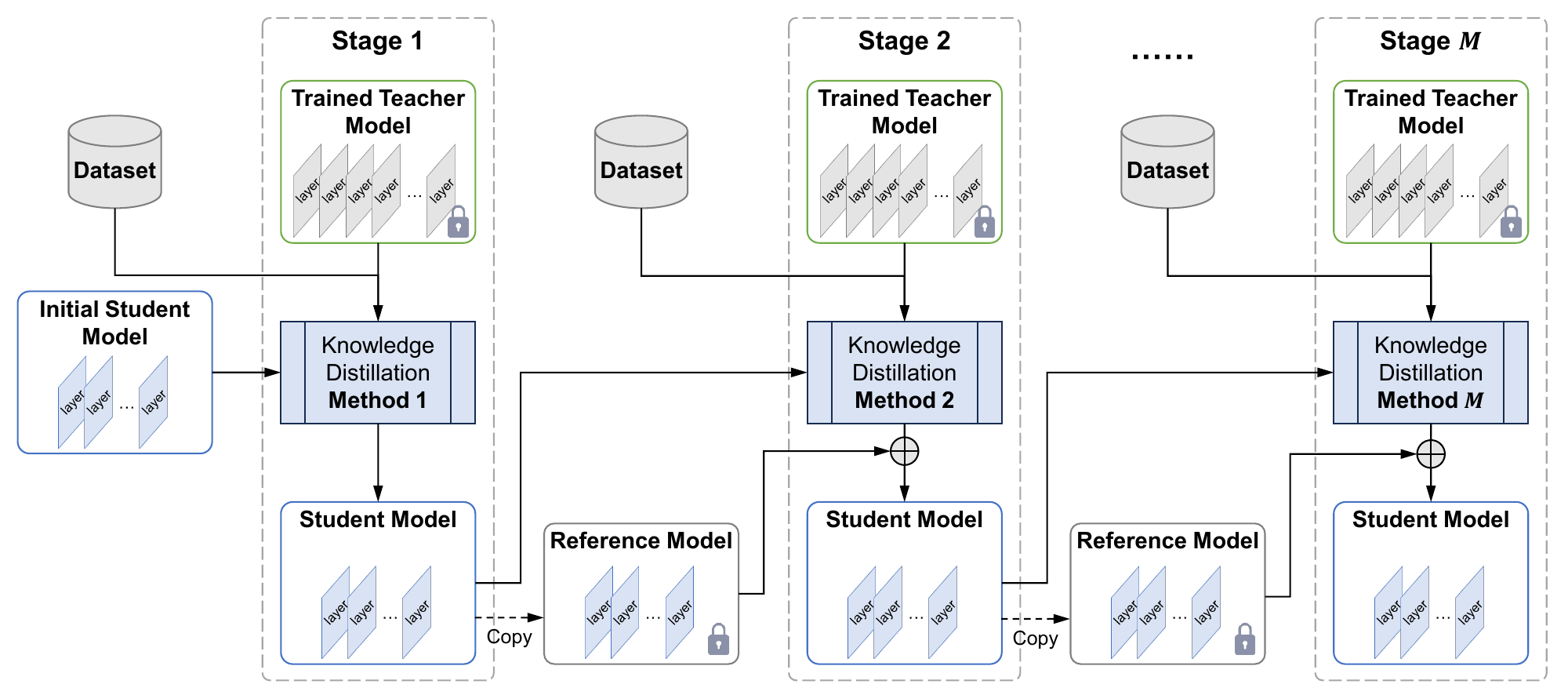}
	\caption{Overview of the Sequential Multi-Stage Knowledge Distillation Framework}
	\label{fig:myarch}
\end{figure}

A schematic overview of the SMSKD workflow is shown in Figure~\ref{fig:myarch} and the detailed steps are summarized in Algorithm~\ref{alg:stagewise}. 
The SMSKD framework organizes the training of the student model into multiple consecutive stages, each characterized by a specific KD method and learning objective. Given the training dataset $\mathcal{D} = \left\{ (\mathbf{x}_i, y_i) \right\}_{i=1}^{N}$, the teacher model $f^T$ pre-trained on $\mathcal{D}$, and the student model $f^S$ to be trained, the SMSKD framework proceeds through $M$ sequential stages, each characterized by a chosen KD method $\text{KD}_m$ and a specified number of training epochs $T_m$. The detailed stage-wise distillation process of SMSKD is described as follows.

%\vspace{-0.5cm}
\begin{algorithm}[htbp]
	\caption{Sequential Multi-Stage Knowledge Distillation}
	\label{alg:stagewise}
	\begin{algorithmic}[1]
		\REQUIRE Training dataset $\mathcal{D} = \left\{ (\mathbf{x}_i, y_i) \right\}_{i=1}^{N}$, teacher model $f^T$ trained on $\mathcal{D}$, student model $f^S$, total number of KD stages $M$,  corresponding KD methods adopted in each stage $\{\text{KD}_1, \cdots, \text{KD}_M\}$, duration (epochs) for each stage $\{T_1, \cdots, T_M\}$, batch size $b$, classification loss weight $\lambda_{c}$, reference loss weight $\lambda_{r}$.
		\STATE Initialize the student model $f^S$.
		\FOR{$m=1,\cdots, M$}
		\FOR{$t=1, \cdots, T_m$}
		\FOR{Batch $B$ in $\text{DataLoader}(\mathcal{D})$}
		\FOR{$(\mathbf{x}_i, y_i) \in B$}
		\IF{$m=1$}
		\STATE Compute loss $\mathcal{L}(\mathbf{x}_i)$ on $\mathbf{x}_i$ with Eq. (\ref{eq:inital}).
		\ELSE
		\STATE Compute loss $\mathcal{L}(\mathbf{x}_i)$ on $\mathbf{x}_i$ with Eq. (\ref{eq:subsequent}).
		\ENDIF
		\ENDFOR
		\STATE $\mathcal{L}_{\text{batch}}=\frac{1}{\left | B \right |}  {\textstyle \sum_{(\mathbf{x}_i, y_i) \in B}}{\mathcal{L}(\mathbf{x}_i)}$.
		\STATE Update the parameters of $f^S$ using $\nabla \mathcal{L}_\text{batch}$.
		\ENDFOR
        \ENDFOR
		\STATE Update the reference model $f^{R} \gets f^S $.
		\ENDFOR
		\ENSURE Student model $f^S$.
	\end{algorithmic}
\end{algorithm}
\vspace{-1cm}

\subsection{Initial Stage}

In the first stage (i.e., $m=1$), the student model $f^S$ is trained using a conventional KD method (e.g., response-based, feature-based, relation-based, or other multi-source KD methods), optimizing a composite loss function $\mathcal{L}_{\text{Stage}}$ that combines the distillation loss with the standard cross-entropy classification loss:
\begin{equation}
\label{eq:inital}
\mathcal{L}_{\text{Stage}} = \mathcal{L}_{\text{Distill}} + \lambda_{c} \cdot \mathcal{L}_{\text{Cls}},
\end{equation}
where $\mathcal{L}_{\text{Distill}}$ denotes the distillation loss specific to the chosen method (that is, for instance, if the vanilla KD method is used, $\mathcal{L}_{\text{Distill}}$ is the loss function defined in Eq.~\ref{eq:vanillaKD}), $\mathcal{L}_{\text{Cls}}$ is the standard classification loss (typically the cross-entropy between the predicted and ground-truth labels), and $\lambda_{c}$ is the weight for the classification loss.

After completing the first stage, the trained student model $f^S$ is copied and frozen as the reference model $f^R$, which will be used in the next stage to anchor subsequent learning.

\subsection{Subsequent Stages}

In each subsequent stage ($m=2, \cdots, M$), the student model is continuously trained using a potentially different KD method. To mitigate catastrophic forgetting of knowledge acquired in previous stages, an adaptive weighted reference loss term $\mathcal{L}_{\text{AdaRef}}$ is introduced, leveraging the frozen reference model from the preceding stage. The total loss $\mathcal{L}_{\text{Stage}}$ for each subsequent stage is defined as:
\begin{equation}
\label{eq:subsequent}
\mathcal{L}_{\text{Stage}} = \mathcal{L}_{\text{Distill}} + \lambda_{c}  \cdot \mathcal{L}_{\text{Cls}} + \lambda_{r}  \cdot \mathcal{L}_{\text{AdaRef}},
\end{equation}
where $\mathcal{L}_{\text{Distill}}$ denotes the distillation loss specific to the chosen method, $\mathcal{L}_{\text{Cls}}$ is the standard classification loss, $\lambda_{c}$ is the weight for the classification loss, $\mathcal{L}_{\text{AdaRef}}$ is the newly introduced adaptive weighted reference loss that is designed to prevent forgetting of knowledge acquired in previous stages, and $\lambda_{r}$ controls the strength of the reference supervision. 

%As different KD methods can be used in each stage and the learning behavior of each method varies, while new knowledge is acquired from the teacher in the current stage, it is possible for the student to forget knowledge learned in previous stages. Thus, mitigating knowledge forgetting or preserving previously acquired knowledge is crucial in our framework and in other sequential learning settings.

As different KD methods may be employed across stages and their learning behaviors differ, acquiring new knowledge from the teacher in the current stage may lead the student to forget knowledge learned from prior stages. Thus, mitigating knowledge forgetting or preserving previously acquired knowledge is crucial in our framework and in other sequential learning settings.

To address this issue, the reference model, inspired by the use of fixed baselines in reinforcement learning (e.g., Proximal Policy Optimization~\cite{schulman2017ppo}), is introduced in our SMSKD framework and acts as a stabilizing anchor that constrains the optimization trajectory of the student model. By incorporating a reference loss, SMSKD encourages the student to remain close to its previous state, thereby preserving useful knowledge and preventing abrupt forgetting when switching distillation strategies. The reference loss $\mathcal{L}_{\text{Ref}}$ can be computed as the Kullback–Leibler (KL) divergence between the output distributions of the student and the reference model,
\begin{equation}
	\label{eq:ref_loss}
	\mathcal{L}_{\text{Ref}} =  \mathrm{KL}\!\left( \mathbf{p}^S \;\|\; \mathbf{p}^R \right),
\end{equation}
where $\mathbf{p}^S$ and $\mathbf{p}^R$ are the output distributions of the student model $f^S$ and reference model $f^R$, respectively. 

However, the reference model $f^R$ may not always provide perfectly accurate or reliable guidance, especially in the early stages. SMSKD employs the true class probability (TCP)~\cite{corbiere2021confidence} as a confidence-based weighting mechanism. For each sample $\mathbf{x}$, the TCP is defined as the reference model's predicted probability for the ground-truth class. Higher TCP values indicate greater confidence, and thus the reference supervision is emphasized for such samples. This adaptive weighting reduces the risk of transferring and preserving erroneous knowledge from the reference model. Thus, the reference loss adaptively weighted by the reference model's confidence (i.e., TCP) is defined as,
\begin{equation}
	\label{eq:adaref_loss}
	\mathcal{L}_{\text{AdaRef}} = \text{TCP} \cdot \mathcal{L}_{\text{Ref}} = p_{k=y}^{R} \cdot \mathrm{KL}\!\left( \mathbf{p}^S \;\|\; \mathbf{p}^R \right),
\end{equation}
where $p_{k=y}^{R}$ is the reference model's predicted probability for the ground-truth class $y$ for sample $\mathbf{x}$.  

At the end of each stage, the updated student model is again copied and frozen as the new reference model for the next stage.
After completing all training stages, the final student model is obtained as the output. By decomposing the distillation process into multiple sequential stages and leveraging both diverse KD methods and reference model anchoring, SMSKD enables the student model to progressively accumulate, retain, and integrate complementary knowledge. The adaptive reference loss further enhances stability and mitigates catastrophic forgetting, resulting in improved performance and robustness.

\section{Experimental Study}
\label{sec:exp_study}

This section comprehensively evaluates the effectiveness of SMSKD framework. The experiments begins with extensive comparisons against state-of-the-art multi-source KD methods, particularly SAKD and Direct Loss Aggregation (DLA), to demonstrate the superiority of our sequential approach. We then explore SMSKD's flexibility by integrating diverse combinations of KD methods that existing approaches cannot handle, validating its constraint-free design. Through detailed ablation studies, we analyze the contribution of each component in our framework, including the reference loss mechanism and the sequential training strategy. Additionally, we investigate the robustness of SMSKD through hyperparameter sensitivity analysis and examine how the number of stages affects its performance. All experiments are conducted on an NVIDIA A100 system with PyTorch. To ensure statistical reliability, each experiment is repeated three times with different random seeds, and the average results are reported.

\subsection{Experimental Setting}

\subsubsection{Datasets}

We conduct experiments on two widely used image classification benchmarks: CIFAR-100\cite{krizhevsky2009learning} and Tiny ImageNet\cite{tinyimagenet}. {CIFAR-100} consists of 60,000 32×32 color images distributed across 100 classes, with 600 images per class. The dataset is split into 50,000 training images and 10,000 test images. Following standard practice, we apply data augmentation: 4-pixel zero-padding, random 32×32 cropping, and random horizontal flipping. Tiny ImageNet is a subset of ImageNet containing 200 classes, with 500 training, 50 validation, and 50 test images per class. All images are resized to 64×64 pixels, and we apply only random horizontal flipping for augmentation. These preprocessing strategies align with prior works~\cite{tian2020contrastive,song2022spot} to ensure fair comparison while preventing overfitting.

\subsubsection{Model Architectures}

To comprehensively evaluate our method, we employ three different teacher-student architecture pairs covering various network families: 1) \textbf{ResNet family}~\cite{he2016deep}: ResNet56 (teacher) → ResNet20 (student); 2) \textbf{WideResNet (WRN) family}~\cite{zagoruyko2016wide}: WRN40\_2 (teacher) → WRN16\_2 (student); and  3) \textbf{Visual Geometry Group (VGG) family}~\cite{simonyan2014very}: VGG13 (teacher) → VGG8 (student). These pairs span varying compression ratios and architectures, enabling a robust assessment of our method's generalizability.

\subsubsection{Compared Methods} 

% We carefully select eight representative distillation methods covering diverse knowledge types to be integrated into our framework: 
% \begin{itemize}
%     \item Response-based method: vanilla KD~\cite{hinton2015distilling} (denoted as KD hereafter for simplicity);
%     \item Feature-based methods: AT (Attention Transfer)~\cite{zagoruyko2017paying}, FitNets~\cite{romero2015fitnets}, and VID (Variational Information Distillation)~\cite{ahn2019variational};
%     \item Relation-based methods: RKD (Relational Knowledge Distillation)~\cite{park2019relational}, CC (Correlation Congruence)~\cite{peng2019correlation}, CRD (Contrastive Representation Distillation)~\cite{tian2020contrastive}, and PKT (Probabilistic Knowledge Transfer)~\cite{passalis2020probabilistic}.
% \end{itemize}

% For comparing with multi-source or integrated KD methods, we consider two baseline approaches:
% \begin{itemize}
%     \item \textbf{SAKD}~\cite{song2022spot}: A state-of-the-art adaptive knowledge integration framework that dynamically selects distillation positions during training. It represents the current best practice in multi-source knowledge integration.
%     \item \textbf{Direct Loss Aggregation (DLA)}: A straightforward baseline that directly sums loss terms from different distillation methods and jointly optimizes them in a single training stage, representing the common practice in prior works without coordination mechanisms.
% \end{itemize}

We carefully select eight representative distillation methods covering diverse knowledge types to be integrated into our framework: (1) Response-based method: vanilla KD~\cite{hinton2015distilling} (denoted as KD hereafter for simplicity); (2) Feature-based methods: AT (Attention Transfer)~\cite{zagoruyko2017paying}, FitNets~\cite{romero2015fitnets}, and VID (Variational Information Distillation)~\cite{ahn2019variational}; and (3) Relation-based methods: RKD (Relational Knowledge Distillation)~\cite{park2019relational}, CC (Correlation Congruence)~\cite{peng2019correlation}, CRD (Contrastive Representation Distillation)~\cite{tian2020contrastive}, and PKT (Probabilistic Knowledge Transfer)~\cite{passalis2020probabilistic}.

For comparing with multi-source or integrated KD methods, we consider two baseline approaches: (1) \textbf{SAKD}~\cite{song2022spot}: A state-of-the-art adaptive knowledge integration framework that dynamically selects distillation positions during training. It represents the current best practice in multi-source knowledge integration; and (2) \textbf{Direct Loss Aggregation (DLA)}: A straightforward baseline that directly sums loss terms from different distillation methods and jointly optimizes them in a single training stage, representing the common practice in prior works without coordination mechanisms.

\subsubsection{Implementation Details}

To ensure fair comparison, we strictly follow the training protocol established by CRD~\cite{tian2020contrastive} and adopted by SAKD~\cite{song2022spot}. All models are trained for 240 epochs using SGD with momentum 0.9, batch size 64, and initial learning rate 0.05. The learning rate is decayed by a factor of 10 at epochs 150, 180, and 210 during the total 240 training epochs. Weight decay is set to 5×10$^{-4}$, and the distillation temperature $\tau$ is fixed at 4.

For individual KD methods, we use their default hyperparameters $\lambda_{c}$ as specified in~\cite{tian2020contrastive}. For our SMSKD framework, we primarily evaluate a two-stage configuration with the following settings: reference loss weight $\lambda_r = 0.5$, classification loss weights $\lambda_c$ consistent with individual method settings, stage 1 duration $T_1=150$ epochs (until the first learning rate decay), and stage 2 duration $T_2=90$ epochs (resulting in a total of $T_1+T_2=240$ epochs). The rationale for these hyperparameter choices is validated through ablation studies and sensitivity analyses presented in subsequent subsections.

\subsection{Comparative Experiments}
\label{sec:comp}

\subsubsection{Comparison with Multi-Source Distillation Methods}
\label{sec:com_integratedKd}
We evaluate the effectiveness of our proposed integration strategy in comparison with SAKD~\cite{song2022spot} and DLA approach. Experiments are conducted on the CIFAR-100 and Tiny ImageNet using three teacher-student architectures: ResNet56$\rightarrow$ResNet20 and WRN40\_2$\rightarrow$WRN16\_2, ResNet56$\rightarrow$ResNet20, and VGG13$\rightarrow$VGG8. 

\begin{table}[hbtp]
\centering
\caption{Test accuracy (\%) of student models on CIFAR-100. 
\textit{Standard} denotes the original accuracy of the corresponding method. For multi-source methods (DLA, SAKD, SMSKD), each integrates the Stage 1 distillation method (listed in the first column) with vanilla KD (Stage 2), and final student's accuracy is reported. Values in parentheses denote the accuracy improvements over the original model.}
\label{tab:cifar_combined}
\resizebox{\linewidth}{!}{%
\begin{tabular}{l|c|ccc|c|ccc}

\hline
\multirow{2}{*}{Method} & \multicolumn{4}{c|}{ResNet56 $\rightarrow$ ResNet20} & \multicolumn{4}{c}{WRN40\_2 $\rightarrow$ WRN16\_2} \\
\cline{2-9}
 & Standard & DLA & SAKD & SMSKD & Standard & DLA & SAKD & SMSKD \\
\hline
Teacher & 73.29 & - & - & - & 75.61 & - & - & - \\
Student & 69.15 & - & - & - & 71.98 & - & - & - \\
KD      & 70.35 & - & - & - & 74.18 & - & - & - \\
\hline
FitNets & 69.42 & 70.68 (+1.26) & 71.45 (+2.03) & \textbf{71.74} (+2.32) 
        & 73.96 & 75.27 (+1.31) & 75.37 (+1.41) & \textbf{75.61} (+1.65) \\
AT      & 70.36 & 71.42 (+0.79) & 71.50 (+0.87) & \textbf{72.16} (+1.53)
        & 74.38 & 75.53 (+1.15) & 75.41 (+1.03) & \textbf{75.97} (+1.59) \\
VID     & 70.38 & 71.00 (+0.62) & 71.36 (+0.98) & \textbf{72.03} (+1.65)
        & 74.47 & 75.05 (+0.58) & 74.93 (+0.46) & \textbf{75.98} (+1.51) \\
CC      & 69.82 & 70.59 (+0.77) & \textbf{71.84} (+2.02) & 71.53 (+1.71)
        & 73.89 & \textbf{75.67} (+1.78) & 75.31 (+1.42) & 75.61 (+1.72) \\
RKD     & 69.85 & \textbf{71.80} (+1.95) & 71.29 (+1.47) & 71.12 (+1.27)
        & 73.45 & \textbf{75.33} (+1.88) & 75.18 (+1.73) & 74.99 (+1.54) \\
CRD     & 71.89 & 71.90 (+0.01) & 71.98 (+0.09) & \textbf{72.42} (+0.53)
        & 75.49 & 75.33 (-0.16) & 75.18 (-0.31) & \textbf{76.02} (+0.53) \\
PKT     & 71.03 & 71.54 (+0.51) & 71.69 (+0.66) & \textbf{71.87} (+0.84)
        & 74.80 & 74.81 (+0.01) & 74.98 (+0.18) & \textbf{75.09} (+0.29) \\
\hline
\end{tabular}%
}
\end{table}

\begin{table}[h]
\centering
\caption{Test accuracy (\%) of student models on Tiny ImageNet. \textit{Standard} denotes the original accuracy of the corresponding method. For multi-source methods (DLA, SAKD, SMSKD), each integrates the Stage 1 distillation method (listed in the first column) with vanilla KD (Stage 2), and final student's accuracy is reported. Values in parentheses denote the accuracy improvements over the original model.}
\label{table:tiny_combined}
\resizebox{\linewidth}{!}{%
\begin{tabular}{l|c|ccc|c|ccc}
\hline
\multirow{2}{*}{Method} & \multicolumn{4}{c|}{ResNet56 $\rightarrow$ ResNet20} & \multicolumn{4}{c}{VGG13 $\rightarrow$ VGG8} \\
\cline{2-9}
 & Standard & DLA & SAKD & SMSKD & Standard & DLA & SAKD & SMSKD \\
\hline
Teacher & 56.51 & - & - & - & 60.38 & - & - & - \\
Student & 51.12 & - & - & - & 55.06 & - & - & - \\
KD      & 53.19 & - & - & - & 58.27 & - & - & - \\
\hline
FitNets & 50.95 & 53.13 (+2.18) & \textbf{54.37} (+3.42) & 53.54 (+2.95) & 56.38 & 59.09 (+2.71) & 58.65 (+2.27) & \textbf{59.18} (+2.80) \\
AT      & 53.67 & 54.39 (+0.72) & 54.35 (+0.68) & \textbf{54.91} (+1.24) & 56.39 & 58.83 (+2.44) & 58.90 (+2.51) & \textbf{59.19} (+2.80) \\
VID     & 52.80 & 53.58 (+0.78) & 53.65 (+0.85) & \textbf{53.91} (+1.11) & 55.93 & 58.51 (+2.58) & 58.48 (+2.55) & \textbf{58.68} (+2.75) \\
CC      & 51.93 & \textbf{54.55} (+2.62) & 54.20 (+2.27) & 53.36 (+1.43) & 55.89 & 58.40 (+2.51) & 58.39 (+2.50) & \textbf{58.42} (+2.53) \\
RKD     & 52.85 & \textbf{54.30} (+1.45) & 54.04 (+1.19) & 53.03 (+0.18) & 56.65 & 58.57 (+1.92) & 58.66 (+2.01) & \textbf{58.71} (+2.06) \\
CRD     & 54.61 & 55.55 (+0.94) & 55.33 (+0.72) & \textbf{55.83} (+1.22) & 57.29 & 59.68 (+2.39) & \textbf{60.25} (+2.96) & 60.19 (+2.90) \\
PKT     & 52.85 & \textbf{54.77} (+1.92) & 54.39 (+1.54) & 53.80 (+0.95) & 57.28 & 58.62 (+1.34) & 58.65 (+1.37) & \textbf{59.78} (+2.50) \\
\hline
\end{tabular}%
}
\end{table}

Table~\ref{tab:cifar_combined} present the results on CIFAR-100 using ResNet56$\rightarrow$ResNet20 and WRN40\_2$\rightarrow$WRN16\_2 architectures, respectively. Table~\ref{table:tiny_combined} reports the results on Tiny ImageNet using ResNet56$\rightarrow$ResNet20 and Vgg13$\rightarrow$Vgg8 architectures, respectively. 
%For each of the seven representative distillation methods, we report (i) the accuracy of the original method alone, (ii) the accuracy after integrating it with KD via three different strategies (our method, SAKD, and DLA), and (iii) the corresponding accuracy gain over the original method.
From the results, it can be observed that all three integration strategies consistently outperform their corresponding single-type distillation baselines across all architectures. This finding indicates that combining multiple distillation objectives yields greater effectiveness than relying on a single one. Notably, our proposed SMSKD method consistently delivers competitive or superior performance across different architectures. In most cases on CIFAR-100, it achieves the highest performance gains among the three integration strategies, with variations observed in CC and RKD. SAKD performs better than DLA in most cases except for RKD, but still underperforms compared to our method overall. Under the CIFAR-100 WRN40\_2$\rightarrow$WRN16\_2 setting, DLA and SAKD sometimes lead to fluctuating or even negative transfer (e.g., CRD degrading from 75.49 to 75.33/75.18), while SMSKD consistently improves performance for all base methods within this configuration. On Tiny ImageNet, while DLA occasionally shows stronger improvements on  the ResNet56$\rightarrow$ResNet20 architecture, SMSKD remains one of the most stable and effective choices. In particular, for the VGG-based architecture, SMSKD shows a clear advantage: it attains the best accuracy in 6 out of 7 distillation methods, with only a negligible gap under CRD (0.06\%). 

The comprehensive comparisons with integrated or multi-source distillation methods demonstrate that our proposed SMSKD approach achieves overall superior performance compared to the baseline methods. We attribute this superiority to several key factors. The DLA method simply aggregates heterogeneous distillation losses without addressing their differing optimization scales, which can negatively impact training dynamics. SAKD improves upon this by enabling layer-wise selection during distillation, but still does not explicitly account for the scale discrepancies among loss terms. In contrast, our SMSKD method mitigates these issues through a multi-stage training scheme that decouples the optimization of different loss scales and leverages a reference model to align the transfer of diverse types of knowledge more effectively and coherently. Importantly, the proposed method achieves these improvements without introducing any additional training cost, demonstrating its efficiency and practical value.

\vspace{-0.4cm}

\subsubsection{Effectiveness on Integrating Various Knowledge Sources}
%In Section~\ref{sec:com_integratedKd}, only limited combinations of KD methods were integrated in our SMSKD framework and compared with SAKD and DLA methods. This limitation arose because SAKD has constraints on the methods that can be integrated—specifically, it only supports combining two methods from different sources, as it leverages multiple knowledge sources including logits and features from the teacher model. In contrast, our proposed SMSKD framework has no such limitations on the methods to be included, since it employs different KD methods in a sequential manner rather than simultaneously. Therefore, in the following experiments, we investigate additional combinations that differ from those examined in Section~\ref{sec:com_integratedKd} to further validate the effectiveness and flexibility of our method.

In previous comparisons, only limited combinations of KD methods were integrated in our SMSKD framework and compared with SAKD and DLA methods, primarily due to SAKD's constraints -- it supports combining only two methods from different sources including logits and features from the teacher model. In contrast, our proposed SMSKD framework has no such limitations on the methods to be included, since it employs different KD methods in a sequential manner rather than simultaneously. Therefore, in the following experiments, we investigate additional combinations that differ from those examined previously to further validate the effectiveness and flexibility of our method.

\begin{figure}[htbp]
\centering
\includegraphics[width=0.8\textwidth]{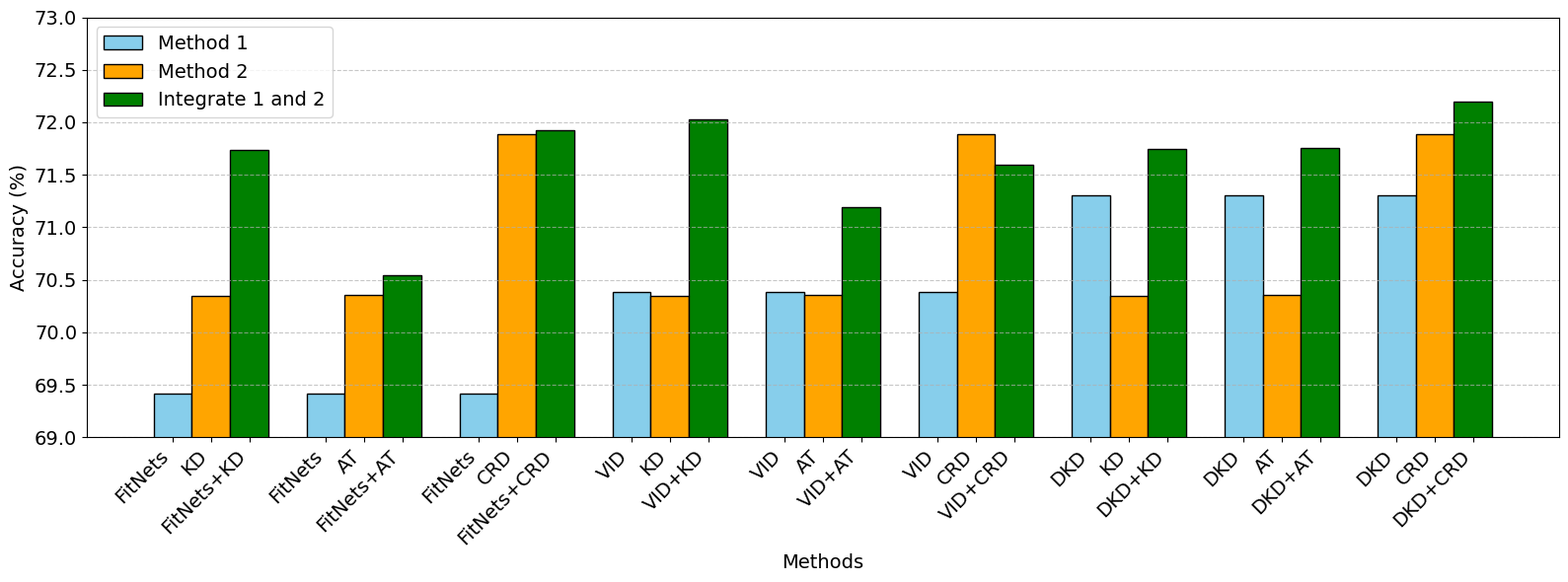}
\caption{Results of integrating more diverse distillation methods within the SMSKD framework on CIFAR-100. Each triplet of bars compares the accuracies of two individual methods (left and middle) and their integration under SMSKD (right), demonstrating its ability to incorporate multiple distillation methods.}
\label{fig:mulmethod}
\end{figure}

Figure~\ref{fig:mulmethod} examines pairwise integrations under SMSKD on CIFAR-100. Each group compares the baselines of two individual methods with their SMSKD-integrated result. Across most combinations, SMSKD consistently outperforms the baselines, demonstrating that the stage-wise strategy effectively exploits complementary knowledge from different distillation objectives. 
A rare exception occurs in the combination of CRD and VID, where performance falls slightly below that of CRD alone. The likely cause is a conflict between CRD’s contrastive objective, which disperses features, and VID’s tendency to pull them toward the teacher distribution. A deeper investigation into such interactions is left for future work.

% A rare exception occurs in the combination of CRD and VID, where the integrated result is slightly worse than using CRD alone. One possible reason is that CRD’s contrastive objective encourages a dispersed instance-level structure in the feature space, 

% which can conflict with VID’s probabilistic feature modeling that partially pulls features toward the teacher’s distribution, thereby weakening the structural separation established by CRD. A deeper investigation into such interactions is left for future work, as understanding such interactions could improve the robustness of multi-stage distillation designs.

%further explores the performance of our SMSKD framework when integrating various KD methods in a pairwise manner. Each group in the figure compares the baseline accuracies of two individual distillation methods and their integrated results obtained through SMSKD. Across most combinations, the sequential integration method consistently outperforms the individual baselines, indicating that our stage-wise strategy can effectively leverage complementary knowledge from diverse distillation objectives.

Notably, combining FitNets (feature-based) with KD (response-based) or CRD (relation-based) yields substantial gains of +2.32\% and +2.50\%, respectively, highlighting the benefit of integrating different knowledge types. Similarly, combining VID (feature-based) and KD achieves accuracy of 72.03\%, surpassing their standalone baselines. Interestingly, although DKD is a response-based method, it yields comparable improvements when integrated with either KD (also response-based) or AT (feature-based), suggesting that even methods from the same supervision source can capture complementary knowledge.

\vspace{-0.5cm}
% \vspace{-1cm}

\subsection{Ablation Study}
\label{sec:ablation}
To systematically evaluate the contribution of each component in SMSKD, we conduct ablation experiments under three progressively enhanced settings:
(1) stage-wise distillation alone, 
(2) additionally applying the reference loss $\mathcal{L}_{\text{Ref}}$, and 
(3) further enabling the adaptive weighted reference loss $\mathcal{L}_{\text{AdaRef}}$ based on TCP.
Table~\ref{tab:ablation_ref_tcp} reports results for integrating seven representative distillation methods with KD method, respectively, where values in parentheses indicate the original standalone accuracy of each method. Unless otherwise specified, all ablation experiments are conducted on CIFAR-100 for fine-grained analysis.

\vspace{-0.5cm}
\begin{table}[htbp]
\centering
\caption{Ablation study on the effects of stage-wise distillation, reference loss, and adaptive weighted reference loss. 
Values in parentheses indicate the accuracy of each distillation method applied individually; rows report results after stage-wise integration with KD and subsequent addition of the supervision terms.
Teacher-Student: WRN40\_2 $\rightarrow$ WRN16\_2.}
\label{tab:ablation_ref_tcp}

\small
\resizebox{\textwidth}{!}{
\begin{tabular}{l|ccccccc}
\hline
Stage 1 Method                      & FitNets (73.96)&AT (74.38)&VID (74.47)&CC (73.89)   &RKD (73.45)&CRD (75.49)&PKT (74.80)\\
Stage 2 Method & KD (75.34)
\\ \hline
Stage-Wise Distillation             & 74.97   & 75.30 & 75.42 & 75.32 & 74.66 & 75.11 & 75.50 \\
Stage-Wise Distillation with $\mathcal{L}_\text{Ref}$           & 75.50   & 75.79 & 75.83 & 75.56 & \textbf{75.28} & 75.86 & \textbf{75.66} \\
Stage-Wise Distillation with $\mathcal{L}_\text{AdaRef}$ & \textbf{75.61} & \textbf{75.97} & \textbf{75.98} & \textbf{75.61} & 74.99 & \textbf{76.02} & 75.09 \\ \hline
\end{tabular}
}
\end{table}
\vspace{-1.1cm}

\subsubsection{Effect of Stage-Wise Distillation}
The third row of Table~\ref{tab:ablation_ref_tcp} shows that stage-wise distillation by integrating the corresponding method with KD approach already yields improvements over the standalone performance of each method (e.g., FitNets 73.96 → 74.97, AT 74.38 → 75.30). This indicates that progressively optimizing the student in a stage-wise manner enables the model to leverage complementary knowledge from different distillation methods, even without additional supervision mechanisms.
\vspace{-0.3cm}
\subsubsection{Effect of Reference Loss}
\begin{figure}[htbp]
    \centering
    \begin{minipage}{0.43\textwidth}
        \centering
        \includegraphics[width=\textwidth]{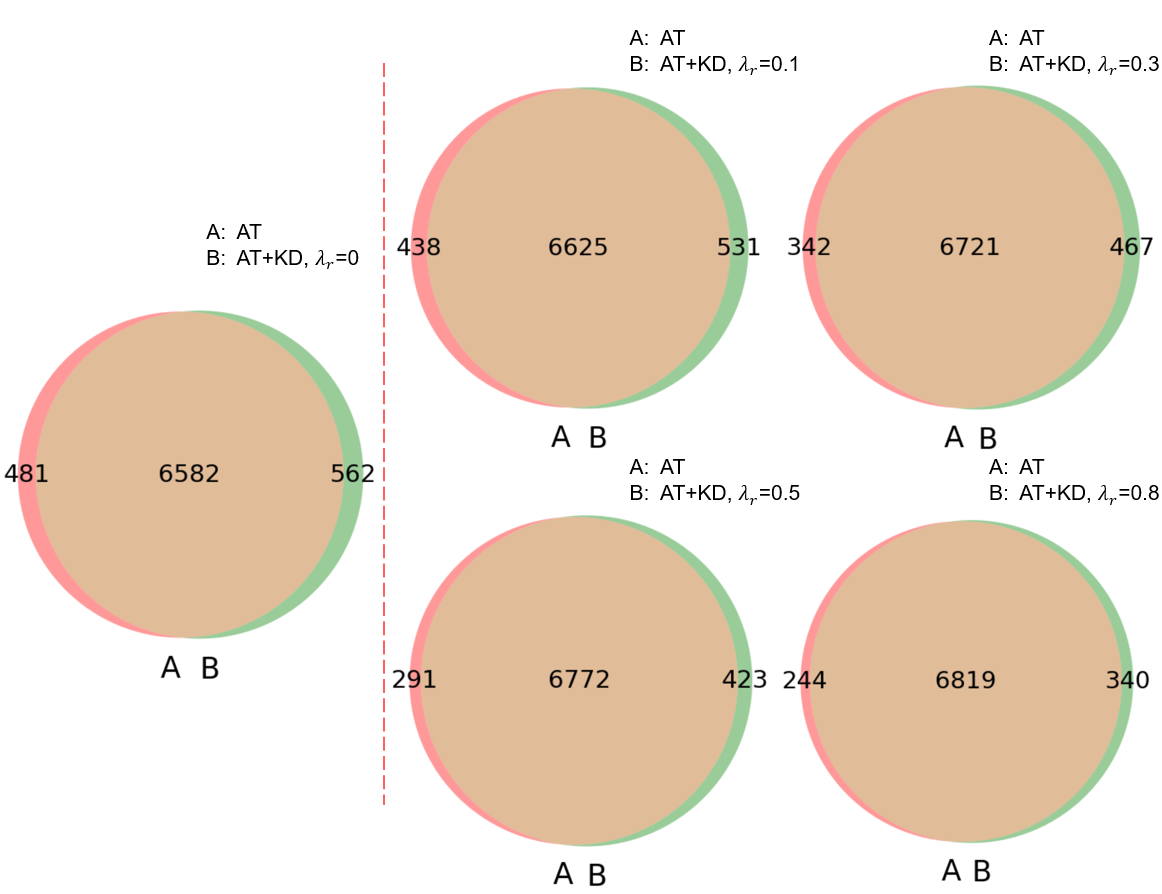}
        \vspace{1mm}
        {\footnotesize (a)}
        \label{fig:venn}
    \end{minipage}
    \hfill
    \begin{minipage}{0.48\textwidth}
        \centering
        \includegraphics[width=\textwidth]{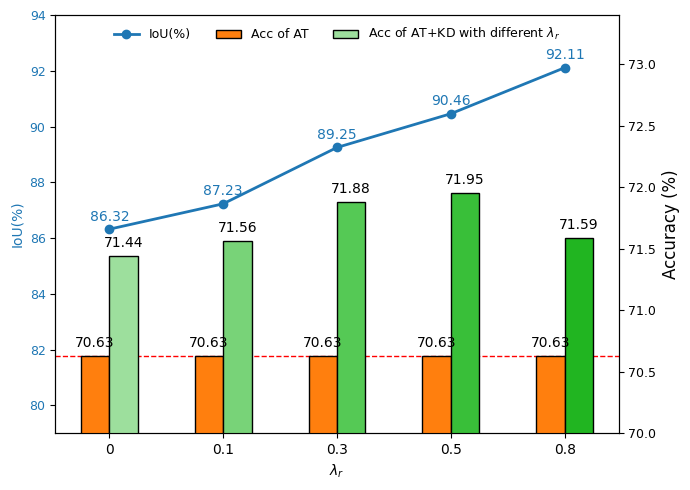}
        \vspace{1mm}
        {\footnotesize (b)}
        \label{fig:reeffect}
    \end{minipage}
\caption{
Effect of the reference loss weight $\lambda_r$ on AT+KD integration for ResNet56$\rightarrow$ResNet20 on CIFAR-100: (a) Venn diagrams (left side: $\lambda_r=0$, without reference supervision; right side: $\lambda_r \in \{0.1,0.3,0.5,0.8\}$, with reference supervision) show overlap of correctly predicted samples between AT-only and AT+KD models. (b) The bar groups depict model accuracy for each $\lambda_r$ value and the blue curve indicates the IoU between correctly predicted sample sets. As $\lambda_r$ increases, the IoU rises, indicating greater overlap in correctly predicted samples.
}
\label{fig:combined_reeffect_venn}
\end{figure}

Building on stage-wise distillation, adding the reference loss $\mathcal{L}_\text{Ref}$ yields stable gains across methods by preventing excessive drift from prior-stage knowledge and mitigating catastrophic forgetting. Figures~\ref{fig:venn} (a) visualizes the effect of the reference loss weight $\lambda_r$ on sample-level agreement for the AT+KD integration via Venn diagrams. Without reference loss ($\lambda_r=0$), 481/6582 (7.3\%) samples correct under AT-only become incorrect under integration (forgotten), while 562 samples (8.5\%) become correct only under integration (newly acquired). Applying reference supervision ($\lambda_r \in {0.1, 0.3, 0.5, 0.8}$) enlarges the union of correctly predicted samples, indicating stronger retention.

Figure~\ref{fig:reeffect}(b) quantifies this by plotting model accuracy (bars) alongside the intersection-over-union (IoU) between correctly predicted sample sets (blue curve) under different $\lambda_r$ values. Specifically, IoU measures the overlap of correct predictions between two models:
$\text{IoU}(A,B)=\frac{|A \cap B|}{|A \cup B|} \times 100\%$,
where $A$ and $B$ are the sets of correctly predicted samples from the two models. 
% A higher IoU indicates greater overlap and better retention of previous knowledge. 
As $\lambda_r$ increases, the IoU rises consistently, indicating greater overlap and better knowledge preservation. However, accuracy gains are most pronounced when $\lambda_r$ increases from 0 to 0.3. Further increasing $\lambda_r$ from 0.5 to 0.8 yields minimal gains and can even degrade performance, suggesting that an excessively strong reference constraint may overly bias the model toward previously learned knowledge while limiting its capacity to acquire new knowledge from the second stage. Thus, the reference loss acts as a balancing mechanism, and a moderate $\lambda_r$ is crucial to maximize both accuracy and cross-stage knowledge retention.

\vspace{-0.5cm}
\vspace{-1.15cm}

\subsubsection{Effect of Adaptive Weighting on Reference Loss}
The last row of Table~\ref{tab:ablation_ref_tcp} shows that incorporating $\mathcal{L}_\text{AdaRef}$ yields additional improvements in most cases, such as FitNets (75.50 → 75.61), AT (75.79 → 75.97), VID (75.83 → 75.98), and CRD (75.86 → 76.02). By adaptively adjusting the reference weight based on each sample's class-prediction confidence (i.e., TCP), the model can more flexibly balance knowledge retention and acquisition. However, slight drops appear in RKD (75.28 → 74.99) and PKT (75.66 → 75.09). This is likely because these relation-based methods emphasize inter-sample structural information, which is less aligned with the class-confidence signal used by TCP for adaptive weighting.

Overall, the ablation study demonstrates that stage-wise distillation and reference loss are the primary factors underlying the observed improvements, while the adaptive weighted term serves as a complementary mechanism that further enhances results in most settings, particularly for response-based and feature-based distillation methods.

\vspace{-0.4cm}
\subsection{Hyperparameters Sensitivity Analysis}

\begin{figure}[htbp]
\centering
\begin{minipage}{0.43\textwidth}
    \includegraphics[width=\textwidth]{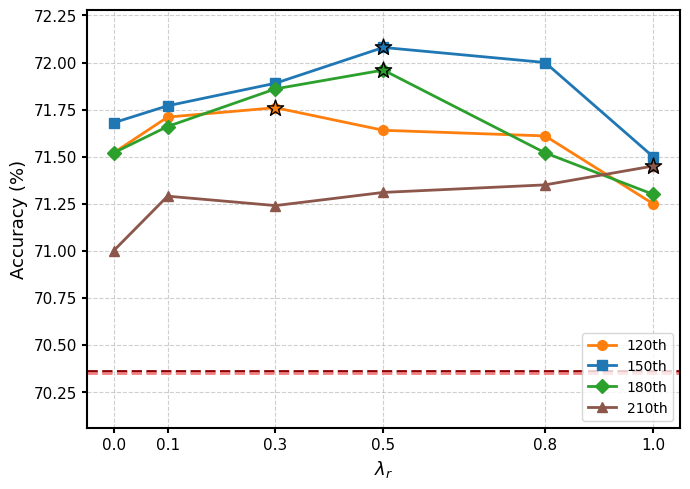}
    \caption{Sensitivity analysis of $\lambda_r$ and the transition epoch on CIFAR-100 using AT+KD integration. Red dashed lines indicate the accuracy of original AT and KD. Teacher-Student: ResNet56-ResNet20.}
    \label{fig:sensitivity}
\end{minipage}
\hfill
\begin{minipage}{0.48\textwidth}
    \includegraphics[width=\textwidth]{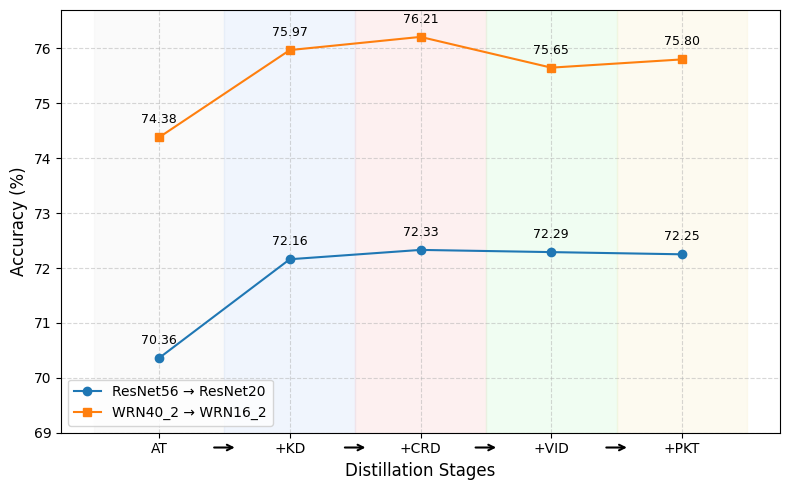}
    \caption{Accuracy of the student model across multiple stages with SMSKD. Experiments are conducted on CIFAR-100 using two teacher-student architectures.}
    \label{fig:fivestage}
\end{minipage}
\end{figure}

Our method introduces two additional hyperparameters: the reference weight $\lambda_{r}$ and the transition epoch. Figure~\ref{fig:sensitivity}  shows how accuracy changes under different settings of these hyperparameters. Specifically, we conduct a grid search with $\lambda_{r}\in \{0.1, 0.3, 0.5, 0.8, 1.0\}$ and transition epoch in $\{120th, 150th, 180th, 210th\}$ to examine their effect. The results indicate that the choices of $\lambda_{r}$ and the transition epoch are not entirely independent. When the transition occurs early, the previous stage may be insufficiently trained, and assigning a large reference weight can mislead the subsequent stage. For example, as shown in Figure~\ref{fig:sensitivity}, when the transition occurs at 120th epoch, the best performance is achieved at $\lambda_{r}=0.3$, whereas at 210 epochs, the optimal $\lambda_{r}$ shifts to 1.0. This finding highlights the practical significance and interpretability of these hyperparameters. Overall, the student model demonstrates robust performance when $\lambda_{r}$ lies between 0.3 and 0.8 and the transition occurs between 120th and 180th epoch. Moreover, across all settings, our proposed framework consistently outperforms the individual methods being integrated (AT: 70.36\%, KD: 70.35\%), indicating that it is not overly sensitive to the precise choice of hyperparameters.

\vspace{-0.5cm}
\vspace{-1.0cm}

\subsection{Scalability Analysis: Beyond Two-Stage Distillation}

We previously employed a two-stage distillation process and found it effective. To assess the scalability of the stage-wise framework, we extend the experiments to three, four, and five stages. Since the original 240 epochs may be insufficient when adding stages, we allocate an extra 90 epochs for each stage beyond the second. For a fair comparison, the minimum learning rate in the extra stage was kept consistent with the previous stages. Specifically, the learning rate at the beginning of the third, fourth and fifth stage was increased from 5e-5 (the final learning rate of the second stage) to 5e-3, and then decayed by a factor of 10 every 30 epochs, ensuring that the final learning rate return to 5e-5.

Figure~\ref{fig:fivestage} reports the student's accuracy across multiple stages using SMSKD. For both architectures, performance improves substantially in the early stages (Stages 1-2) and then stabilizes (ResNet56 $\rightarrow$ ResNet20) or fluctuates slightly (WRN40\_2$\rightarrow$WRN16\_2) in the later stages. This diminishing return pattern suggests that the student model approaches its capacity limit after two or three stages, and additional stages provide limited benefit while increasing training cost (90 epochs per stage). Overall, these results reveal a trade-off between accuracy and efficiency. SMSKD scales well up to two stages, and beyond that, diminishing returns indicate that two stages strike the best balance in most scenarios. Nonetheless, the framework can accommodate more stages when marginal improvements are critical and compute is abundant.

%Future work could explore adaptive stage selection that determines the optimal number of stages based on validation performance, or develop specialized objectives for later stages to mitigate capacity saturation.

\section{Conclusion}
\label{sec:conclusion}

In this work, we propose SMSKD (Sequential Multi-Stage Knowledge Distillation), a simple yet effective framework that sequentially integrates heterogeneous KD methods to achieve superior student performance. Unlike prior approaches combining multiple KD objectives simultaneously, SMSKD adopts a progressive training strategy that allows the student to absorb complementary knowledge from different distillation methods in successive stages. To address catastrophic forgetting during stage transitions, we introduce a reference model supervision mechanism that anchors the student to its previous-stage state, preventing excessive deviation while enabling effective knowledge acquisition. Furthermore, we propose an adaptive weighting strategy based on the teacher's true class probability (TCP) that dynamically adjusts reference loss strength for each sample, providing flexible balance between knowledge retention and integration. 
Extensive experiments demonstrate that SMSKD consistently improves student accuracy across diverse teacher–student architectures and KD method combinations and achieves superior performance compared to existing methods. Ablation studies confirm that stage-wise distillation and reference loss are the primary contributors, while TCP-based adaptive weighting provides complementary benefits in most settings. Notably, SMSKD introduces negligible computational overhead and imposes no constraints on the distillation methods to be integrated, making it a practical and resource-efficient approach for integrating KD methods.

\subsubsection{\ackname} This work was supported by the National Natural Science Foundation of China (Grant No. 62250710682), an internal grant of Lingnan University, the Guangdong Provincial Key Laboratory (Grant No. 2020B121201001), and the Program for Guangdong Introducing Innovative and Entrepreneurial Teams (Grant No. 2017ZT07X386).

\bibliographystyle{splncs03}
\bibliography{ref}

\begin{thebibliography}{10}
\providecommand{\url}[1]{\texttt{#1}}
\providecommand{\urlprefix}{URL }

\bibitem{ahn2019variational}
Ahn, S., Hu, S.X., Damianou, A., Lawrence, N., Dai, Z.: Variational information distillation for knowledge transfer. In: Proceedings of the IEEE/CVF Conference on Computer Vision and Pattern Recognition (CVPR). pp. 9155--9163 (2019)

\bibitem{cheng2018model}
Cheng, Y., Wang, D., Zhou, P., Zhang, T.: Model compression and acceleration for deep neural networks: The principles, progress, and challenges. IEEE Signal Processing Magazine  35(1),  126--136 (2018)

\bibitem{corbiere2021confidence}
Corbière, C., Thome, N., Saporta, A., Vu, T.H., Cord, M., Pérez, P.: Confidence estimation via auxiliary models. IEEE Transactions on Pattern Analysis and Machine Intelligence  44(10),  6043--6055 (2022)

\bibitem{cui2024decoupled}
Cui, J., Tian, Z., Zhong, Z., Qi, X., Yu, B., Zhang, H.: Decoupled kullback-leibler divergence loss. Advances in Neural Information Processing Systems (NeurIPS)  37,  74461--74486 (2024)

\bibitem{gou2021knowledge}
Gou, J., Yu, B., Maybank, S.J., Tao, D.: Knowledge distillation: A survey. International Journal of Computer Vision  129(3),  1789--1819 (2021)

\bibitem{he2016deep}
He, K., Zhang, X., Ren, S., Sun, J.: Deep residual learning for image recognition. In: Proceedings of the IEEE Conference on Computer Vision and Pattern Recognition (CVPR). pp. 770--778 (2016)

\bibitem{hinton2015distilling}
Hinton, G., Vinyals, O., Dean, J.: Distilling the knowledge in a neural network (2015), arXiv:1503.02531

\bibitem{huang2024harmonizingknowledgetransferneural}
Huang, Y., Yan, Z., Shen, C., Fang, F., Zhang, G.: Harmonizing knowledge transfer in neural network with unified distillation. In: European Conference on Computer Vision. pp. 58--74 (2024)

\bibitem{krizhevsky2009learning}
Krizhevsky, A.: Learning multiple layers of features from tiny images. Tech. rep., University of Toronto (2009)

\bibitem{lecun2015deep}
LeCun, Y., Bengio, Y., Hinton, G.: Deep learning. Nature  521(7553),  436--444 (2015)

\bibitem{tinyimagenet}
Li, F.F., Karpathy, A., Johnson, J.: Tiny imagenet visual recognition challenge. \url{https://tinyimagenet.stanford.edu/} (2015)

\bibitem{mansourian2025comprehensivesurveyknowledgedistillation}
Mansourian, A.M., Ahmadi, R., Ghafouri, M., Babaei, A.M., Golezani, E.B., Ghamchi, Z.Y., Ramezanian, V., Taherian, A., Dinashi, K., Miri, A., Kasaei, S.: A comprehensive survey on knowledge distillation (2025), arXiv:2503.12067

\bibitem{miles2024vkd}
Miles, R., Elezi, I., Deng, J.: \textit{V}\textsubscript{k}\textit{D}: Improving knowledge distillation using orthogonal projections. In: Proceedings of the IEEE/CVF Conference on Computer Vision and Pattern Recognition (CVPR). pp. 15720--15730 (2024)

\bibitem{OjhaLRLL23}
Ojha, U., Li, Y., Sundara~Rajan, A., Liang, Y., Lee, Y.J.: What knowledge gets distilled in knowledge distillation? In: Advances in Neural Information Processing Systems (NeurIPS). vol.~36, pp. 11037--11048 (2023)

\bibitem{park2019relational}
Park, W., Kim, D., Lu, Y., Cho, M.: Relational knowledge distillation. In: Proceedings of the IEEE/CVF Conference on Computer Vision and Pattern Recognition (CVPR). pp. 3967--3976 (2019)

\bibitem{passalis2020probabilistic}
Passalis, N., Tzelepi, M., Tefas, A.: Probabilistic knowledge transfer for lightweight deep representation learning. IEEE Transactions on Neural Networks and learning systems  32(5),  2030--2039 (2020)

\bibitem{peng2019correlation}
Peng, B., Xu, X., Feng, D., Huang, W., Tao, D.: Correlation congruence for knowledge distillation. In: Proceedings of the IEEE/CVF International Conference on Computer Vision (ICCV). pp. 5007--5016 (2019)

\bibitem{romero2015fitnets}
Romero, A., Ballas, N., Kahou, S.E., Chassang, A., Gatta, C., Bengio, Y.: {FitNets}: Hints for thin deep nets. In: Proceedings of the International Conference on Learning Representations (ICLR). San Diego, CA, USA (2015)

\bibitem{schulman2017ppo}
Schulman, J., Wolski, F., Dhariwal, P., Radford, A., Klimov, O.: Proximal policy optimization algorithms (2017), arXiv:1707.06347

\bibitem{shen2019customizing}
Shen, C., Xue, M., Wang, X., Song, J., Sun, L., Song, M.: Customizing student networks from heterogeneous teachers via adaptive knowledge amalgamation. In: Proceedings of the IEEE/CVF International Conference on Computer Vision (ICCV). pp. 3504--3513 (2019)

\bibitem{simonyan2014very}
Simonyan, K., Zisserman, A.: Very deep convolutional networks for large-scale image recognition. arXiv preprint arXiv:1409.1556  (2014)

\bibitem{song2022spot}
Song, J., Chen, Y., Ye, J., Song, M.: Spot-adaptive knowledge distillation. IEEE Transactions on Image Processing  31,  3359--3370 (2022)

\bibitem{tian2020contrastive}
Tian, Y., Krishnan, D., Isola, P.: Contrastive representation distillation. In: Proceedings of the International Conference on Learning Representations (ICLR). Addis Ababa, Ethiopia (2020)

\bibitem{zagoruyko2016wide}
Zagoruyko, S., Komodakis, N.: Wide residual networks. In: Proceedings of the British Machine Vision Conference (BMVC). pp. 87.1--87.12 (2016)

\bibitem{zagoruyko2017paying}
Zagoruyko, S., Komodakis, N.: Paying more attention to attention: Improving the performance of convolutional neural networks via attention transfer. In: Proceedings of the International Conference on Learning Representations (ICLR). Toulon, France (2017)

\bibitem{zhao2022decoupled}
Zhao, B., Cui, Q., Song, R., Qiu, Y., Liang, J.: Decoupled knowledge distillation. In: Proceedings of the IEEE/CVF Conference on Computer Vision and Pattern Recognition (CVPR). pp. 11928--11937 (2022)

\bibitem{zheng2023rotatedLD}
Zheng, Z., Ye, R., Hou, Q., Ren, D., Wang, P., Zuo, W., Cheng, M.M.: Localization distillation for object detection. IEEE Transactions on Pattern Analysis and Machine Intelligence  45(8),  10070--10083 (2023)

\end{thebibliography}

\end{document}